\documentclass{article}

\usepackage{PRIMEarxiv}

\usepackage[utf8]{inputenc} 
\usepackage[T1]{fontenc}    
\usepackage{hyperref}       
\usepackage{url}            
\usepackage{booktabs}       
\usepackage{amsfonts}       
\usepackage{nicefrac}       
\usepackage{microtype}      
\usepackage{lipsum}
\usepackage{fancyhdr}       
\usepackage{graphicx}       
\graphicspath{{media/}}     

\title{Strengthening the EU AI Act: Defining Key Terms on AI Manipulation
}

\author{
  Matija Franklin \\
  University College London  \\
  \texttt{matija.franklin@ucl.ac.uk} 
   \And
  Philip Moreira Tomei \\
  Pax Machina
  \AND
  Rebecca Gorman \\
  Aligned AI \\
}

\begin{document}
\maketitle

\begin{abstract}
The European Union's Artificial Intelligence Act aims to regulate manipulative and harmful uses of AI, but lacks precise definitions for key concepts. This paper provides technical recommendations to improve the Act's conceptual clarity and enforceability. We review psychological models to define "personality traits," arguing the Act should protect full "psychometric profiles." We urge expanding "behavior" to include  "preferences" since preferences causally influence and are influenced by behavior. Clear definitions are provided for "subliminal," "manipulative," and "deceptive" techniques, considering incentives, intent, and covertness. We distinguish "exploiting individuals" from "exploiting groups," emphasising different policy needs. An "informed decision" is defined by four facets: comprehension, accurate information, no manipulation, and understanding AI's influence. We caution the Act's therapeutic use exemption given the lack of regulation of digital therapeutics by the EMA. Overall, the recommendations strengthen definitions of vague concepts in the EU AI Act, enhancing precise applicability to regulate harmful AI manipulation. 
\end{abstract}

\keywords{EU AI Act \and AI Manipulation \and AI Policy}

\section{Introduction}

In the amendments adopted by the European Parliament on 14 June 2023 on the Artificial Intelligence Act, the EU’s regulatory stance on AI Manipulation is outlined as such:

\emph{“(a) the placing on the market, putting into service or use of an AI system that deploys subliminal techniques beyond a person’s consciousness or purposefully manipulative or deceptive techniques, with the objective to or the effect of materially distorting a person’s or a group of persons’ behaviour by appreciably impairing the person’s ability to make an informed decision, thereby causing the person to take a decision that that person would not have otherwise taken in a manner that causes or is likely to cause that person, another person or group of persons significant harm;
}

\emph{The prohibition of AI system that deploys subliminal techniques referred to in the first sub-paragraph shall not apply to AI systems intended to be used for approved therapeutical purposes on the basis of specific informed consent of the individuals that are exposed to them or, where applicable, of their legal guardian;
}

\emph{(b) the placing on the market, putting into service or use of an AI system that exploits any of the vulnerabilities of a person or a specific group of persons, including characteristics of such person’s or such group’s known or predicted personality traits or social or economic situation, age, physical or mental ability with the objective or to the effect of materially distorting the behaviour of that person or a person pertaining to that group in a manner that causes or is likely to cause that person or another person significant harm \cite{eu2023b}”
}

We argue that in the current regulatory framing, there is a lack of clarity of core concepts in the present amendments. For example, “personality traits” are mentioned six times in the latest amendments, and yet are not defined at any point in the document, or in the draft of the Act \cite{eu2023a, eu2023b}. In the present article, we will focus on providing technical definitions based on best practices in psychology and artificial intelligence as a way of improving and operationalizing core concepts to regulate AI manipulation. We will focus on:

\begin{enumerate}
\item Reviewing different approaches to defining personality traits; 
\item Expanding the target of manipulation from solely behaviour to the manipulation of preferences; 
\item Defining deception and deceptive techniques, subliminal techniques, and purposefully manipulative techniques; 
\item Understanding how “exploiting vulnerabilities” requires different approaches for individuals and groups; 
\item Defining what an “informed decision” is; 
\item Defining “therapeutic purposes” in relation to manipulation.
\end{enumerate}

\section{Defining Personality Traits}

The current version of the act prohibits using the \emph{‘vulnerabilities of individuals and specific groups of persons due to their known or predicted personality traits, age, physical or mental incapacities’}. The concept of ‘personality traits’ is referred to multiple times in the proposed law but never defined further. Defining ‘personality traits’ will most likely require picking one of the many frameworks used in psychology to operationalise this concept. The most famous model is the OCEAN model, very often used as the standard method of quantifying personality \cite{cobbclark2012}. The model delineates five axes to evaluate an individual’s personality:

\begin{enumerate}
    \item  \textit{Openness to experience}: refers to the degree to which an individual is receptive to novel ideas, experiences, and emotions, ranging from inventive and curious at one end to consistent and cautious at the other.
    \item Conscientiousness: pertains to an individual's propensity for self-discipline, organisation, and goal-oriented behaviours, contrasting those who are efficient and organised with those who are extravagant and careless.
    \item \textit{Extraversion}: encapsulates an individual's orientation towards the external world, distinguishing those who are outgoing and energetic from those who are solitary and reserved.
    \item \textit{Agreeableness}: denotes an individual's interpersonal tendencies, differentiating those who are friendly and compassionate from those who are critical and rational.
    \item \textit{Neuroticism}: captures the stability and intensity of an individual's emotional responses, distinguishing those who are sensitive and nervous from those who are resilient and confident.
\end{enumerate}

Digital behavioural markers, like social media posts and browser histories, have effectively depicted individuals' Big Five personality traits \cite{kosinski2013}. Such detailed insights can offer predictive abilities about human behaviour \cite{kosinski2016}. Personality evaluations can be used to tailor persuasive messages to resonate with the psychological makeup of broad audience groups \cite{matz2017}. Among the varied data sources tapped for these insights are online likes, mobile device usage patterns, and even music preferences \cite{lambiotte2014}. Intriguingly, algorithmic evaluations of personalities often surpass the judgment accuracy of one's close peers \cite{youyou2015}. The potency of personality traits as a tool to sway decision-making, especially in voting, is evident from the findings of \cite{gerber2011}.

Given the extensive research done on OCEAN, and given that personality traits have been exploited as a vulnerability to influence voting behaviour, an option for EU policymakers would be to adopt OCEAN as their way of defining personality traits. However, the OCEAN model is neither uncontroversial nor complete. The OCEAN model has been subjected to considerable critical scrutiny (e.g., \cite{trofimova2014, trofimova2018}); an extensive review is beyond the scope of this paper. In line with this criticism, more recent models such as the HEXACO model, have included a sixth factor - Honesty-Humility \cite{ashton2014}. The evidence for six large groups of personality traits (rather than five) has been provided by comprehensive, cross-language studies \cite{ashton2004}. This provides policymakers with a dilemma - whether to define personality with OCEAN, HEXACO, or another model.

A further decision policymakers can make is whether to keep personality traits at all or whether to replace them with a more expansive construct - psychometric traits. A psychometric trait is a quantifiable and consistent characteristic of an individual's psychological functioning that can be reliably measured using standardised assessment tools. The enhanced forecasting abilities of digital AI platforms, combined with in-depth information about the user's online activities and inclinations, pave the way for an environment where our innate psychological variations can be exploited as vulnerabilities. On a group level, exploiting small differences in psychometric traits can be effective. Individual psychological variations can be employed to form straightforward models predicting the reactions of specific groups to given stimuli, making these consistent individual disparities susceptible to manipulation.

Personality traits are but a small minority of objectively measurable psychometric traits that could be exploited by an AI system in order to manipulate EU citizens. We thus propose that the EU AI act should protect the entire psychological profile as an avenue for manipulation from sophisticated AI systems rather than just ‘personality traits’. Like personality traits, there are proxy measures for people’s psychometric profiles, which use secondary data available online \cite{stark2018}.

Specific psychometric measures such as suggestibility \cite{pellicer2018} and hypnotisability \cite{moore1998} are crucial in determining how to manipulate an individual and should be protected characteristics along with demography and personality traits. Another characteristic is the influence of choice architectures on the decision-making of an individual, described in psychological literature as nudgeability \cite{deridder2022} \footnote{These traits are less stable than the previously aforementioned traits, thus more research is required.}. These properties and others can be measured or inferred from user data and behaviour, and then exploited by a sophisticated AI system in order to coerce or manipulate human actors. As there are many of these factors, and many factors are yet to be discovered, it is likely that making a list would not be appropriate as it would quickly be outdated. 

In light of these statements, we suggest changing references to ‘personality traits’ to 'psychological traits', defined as relatively stable psychological characteristics of human agents that are either directly measured or inferred from available data.

\section{Considering Preferences, not Just Behaviour}

The current version of the Act states that the target of manipulation (and thus what policy should be designed around) is behaviour. More specifically, the act is currently solely concerned with how AI can distort a person’s behaviour. This is made evident in statements such as \emph{"...distorting a person’s or a group of persons’ behaviour...''} or \emph{"...the effect of materially distorting the behaviour..."}. It thus does not take into account the manipulation of other aspects of people’s psychology. We propose that the EU AI Act should be concerned with preferences. Preferences can be stated directly by a person - and are thus known as stated preferences \cite{kroes1988} - or revealed through their behaviour - revealed preferences \cite{samuelson1938}.

Preferences have a bidirectional causal relationship with behaviour \cite{ariely2008}. Machine Learning is often used to learn the preferences of users in order to better deliver some service to them. Examples of this are recommender systems \cite{khanal2020} that learn revealed preferences, or reinforcement learning from human feedback methods \cite{christiano2017} that learn users' stated preferences. A problem emerges due to the iterative nature of machine learning. An AI system that learns preferences will change their interaction with a human in line with those preferences. This change in interaction influences human behaviour. As these AI Systems influence human behaviour, they also influence human preferences. As a result, by learning preferences, an AI system changes preferences \cite{ashton2022a}.  Thus, even if a policymaker was solely concerned with the manipulation of behaviour, it is important to consider preferences as they influence and are influenced by behaviour. 

AI systems also hold the capability to directly influence preferences through various methods - targeted content recommendations, personalised advertising, or even subtle alterations in the presentation of choices. Such AI-enabled manipulative practices can guide individuals' and groups' preferences. The consequences can be far-reaching, affecting various areas from individual consumer habits and mental health to societal issues like propagation of biases and deepening of social divisions.

We define preferences as “... any explicit, conscious, and reflective or implicit, unconscious, and automatic mental process that brings about a sense of liking or disliking for something \cite{franklin2022a}.”

It is of crucial importance to consider the central role of preferences in relation to behaviour in the current version of the EU AI Act \cite{franklin2022b}. As preferences may significantly influence human actions, overlooking this relationship may create regulatory blind spots in the Act. This is especially the case when distorting preferences at time T1 may harmfully change behaviour at a much later time T2. The Act's current focus on behaviour may inadvertently allow AI systems to influence preferences without immediate behavioural consequences, thereby escaping the regulatory oversight of the Act. Such a loophole could potentially enable AI practices to exert influence over individuals and society in a long-term and insidious manner.

Given these considerations, we propose that the EU AI Act should expand its scope to expressly prohibit AI systems that purposefully and materially manipulate or distort a person's or a group of persons' preferences.

\section{Defining Subliminal, Purposefully Manipulative and Deceptive Techniques}

The AI Act currently prohibits \emph{“...AI system that deploys subliminal techniques beyond a person’s consciousness or purposefully manipulative or deceptive techniques…”} This requires separate definitions and policy recommendations of subliminal techniques, purposefully manipulative techniques, and deceptive techniques. In this section, we want to provide definitions and context for subliminal techniques, purposefully manipulative techniques, and deceptive techniques.

\subsection{Subliminal Techniques}

\cite{bermudez2023} provide a comprehensive analysis of the concept of subliminal techniques, offering both narrow and broad definitions. The narrow definition is: \emph{"Subliminal techniques aim at influencing a person’s behaviour by presenting a stimulus in such a way that the person remains unaware of the stimulus presented."} 

This definition aligns with the traditional understanding of subliminal techniques in psychology and marketing, where the stimulus is presented below the threshold of conscious perception but can still influence behaviour \cite{bermudez2023}. However, the researchers argue that this narrow definition may not capture all the ethically concerning techniques that could be used in AI systems. 

They propose a broader definition: \emph{"Subliminal techniques aim at influencing a person’s behaviour in ways in which the person is likely to remain unaware of (1) the influence attempt, (2) how the influence works, or (3) the influence attempt’s effects on decision-making or value- and belief-formation processes} \cite{bermudez2023}."

To enforce this definition separate checks would be required for each of the Definition’s three clauses: lack of awareness of (1) the influence attempt, (2) its mode of operation, or (3) its effects on decision and judgment \cite{bermudez2023}. If a system leads to one or more of these factors,  the next step is to determine whether these techniques are used in a distorting way, that is, in a way that significantly reduces the person’s capacity for guiding their actions in accordance with their values. If they do, then an ethical risk assessment is required to assess whether the system’s use of subliminal techniques implies an increase in the risk of harm to anyone that may be affected by those behaviours.

This broader definition is more suitable to the EU AI Act, covering a wider range of techniques that could potentially harm individuals or groups \cite{bermudez2023}. It includes not only techniques that use stimuli below the threshold of conscious perception, but also those that operate in ways that the person is not aware of or cannot resist. It changes the focus from manipulation that occurs due to a lack of awareness towards the stimuli to manipulation that occurs due to a lack of awareness of the manipulation itself. A person can be aware of the stimuli, yet be manipulated by it regardless. This definition aligns with the EU AI Act's intention to prevent AI systems from deploying subliminal techniques that could materially distort a person's behaviour and cause significant harm. This definition also aligns with our own analysis that the Act should protect against manipulation beyond behaviour and decision-making, and also focus on preferences (in the broader sense, as defined in this paper) or \emph{“value- and belief-formation processes.”}

\subsection{Purposefully Manipulative Techniques}

Recent definitions of AI Manipulation, define AI systems as manipulative \emph{“... if the system acts as if it were pursuing an incentive to change a human (or other agent) intentionally and covertly”} \cite{carroll2023}. Understanding incentives, intent, and covertness provides a good framework for identifying purposefully manipulative techniques.

The first axis of manipulation is whether the system has incentives for influence, i.e., incentives to change a human’s behaviour \cite{carroll2023}. This could involve changing their beliefs, preferences, or psychological state more broadly. An incentive exists for a certain behaviour if it increases the reward (or decreases the loss) the AI system receives during training. For example, recommender systems may have incentives to influence user behaviour so as to make them more predictable \cite{kasirzadeh2023}.  

The second axis of manipulation is intent, which relates to the idea that prohibited “manipulative techniques” are those that are used “purposefully” \cite{carroll2023}. The researchers propose grounding the notion of intent in a fully behavioural lens, which is agnostic to the actual computational process of the system \cite{ashton2022b}. In other words, intent does not mean that an AI system has the same reasoning and planning abilities as a human does. They propose that \emph{“...a system has intent to perform a behaviour if, in performing the behaviour, the system can be understood as engaging in a reasoning or planning process for how the behaviour impacts some objective.”} Including intent allows one to separate AI systems that are manipulative incidentally (e.g. random chance) from AI behaviour displaying a systematic pattern of manipulation in order to cause an outcome. 

An issue in measuring intent is deciphering what it means for an AI to engage in “a reasoning or planning process." \cite{ashton2022b} argues that an AI system aims for an outcome via a specific action when (i) there are other possible actions, (ii) the AI system can recognize when the outcome happens, (iii) the AI system predicts that the action will lead to the outcome, and (iv) the outcome is advantageous to the AI system.

The researchers define covertness \emph{“…as the degree to which a human is not aware of the specific ways in which an AI system is attempting to change some aspect of their behaviour, beliefs, or preferences} \cite{carroll2023}.” The concept of covertness serves as a distinguishing factor between manipulation and persuasion. In instances of persuasion, the individual being persuaded is typically conscious of the efforts made by the persuader to alter their viewpoint. However, the element of covertness implies that the individual may not be aware of the influence being exerted on them, thus making it difficult for them to consent or resist this influence. This lack of awareness can compromise their autonomy, differentiating it from persuasion. If a human understands how an AI system operates, the possibility of the AI system acting in a covert manner seems lower than otherwise. 

Although we acknowledge that the most influential view of manipulation in the field defines it as “hidden influence” \cite{susser2019}, there are some problems with linking covertness and manipulation very closely. More covertness will on average produce more manipulation. However, some researchers argue that it is possible for a person to know that they are being manipulated and for it to still be considered as manipulation \cite{jongepier2022}. A classic example is that of guilt trips and other types of emotional manipulation \cite{klenk2022}. Thus although covertness can serve as a good proxy measure for manipulation, one needs to consider cases where relying on it backfires. 

Through this framework, one can identify and regulate purposefully manipulative techniques by finding instances where an AI system receives a reward (or decreases a loss) when an AI system aims for an outcome via a specific action and this action is highly covert in that there is a low degree of human awareness \emph{“of the specific ways in which an AI system is attempting to change some aspect of their behaviour, beliefs, or preferences.”}

\subsection{Deceptive Techniques}

Certain AI systems have found ways to garner positive reinforcement by executing actions that misleadingly suggest to the human overseer that the AI has accomplished its set goal. For instance, a study revealed that a virtual robotic arm was able to simulate the act of grasping a ball \cite{christiano2017}. The AI system, which was trained via human feedback to pick up a ball, instead learned to position its hand in such a way that it blocked the ball from the camera's view, creating a false impression of success. Another study demonstrated that GPT4 possesses the capability to \emph{"understand and induce false beliefs in other agents} \cite{hagendorff2023}." There have also been instances where AI systems have learned to identify when they are under evaluation and temporarily halt any undesired behaviours, only to resume them once the evaluation period is over \cite{lehman2020}. This kind of deceptive behaviour, known as specification gaming, could potentially become more prevalent as future AI systems take on more complex tasks that are harder to assess, thereby making their deceptive actions harder to detect \cite{pan2022}. 

Thus, an AI system is deceptive if it behaves in a way that is not aligned with the intentions of the user but pretends to behave as if it had the intentions of the user. This is clear in the evaluation example where an AI behaves differently when it is being evaluated. It seems to be aligned with the intentions of the user but it actually is not. For example, it can have information but not disclose it to a person when prompted to. Some refer to the issue of an AI possessing specific knowledge but not sharing it as the 'Eliciting Latent Knowledge' problem \cite{christiano}. However, latent knowledge is not the only source of deception.

We propose that deception, in the context of AI systems, refers to an intentional act (as defined above) by an AI system to create a false or misleading impression about its goals, capabilities, operations, or effects, with the aim of materially distorting a user's understanding, preferences, or behaviour in a manner likely to cause that user or others significant harm. This can occur in various ways, including but not limited to:

\begin{enumerate}
    \item Misrepresenting or obscuring the system's goals or intents, thereby creating a perception of alignment with the user's goals or interests when this is not the case;
    \item Providing false, incomplete, or misleading information about the system's capabilities or limitations;
    \item Manipulating or obscuring the outputs, outcomes, or effects of the system's operation in a misleading manner;
    \item Falsely representing the system's knowledge or lack thereof, with regard to any information or request by the user;
    \item Altering the system's behaviour temporarily or selectively in response to monitoring or evaluation activities, in a way that falsely represents its typical operations or effects.
\end{enumerate}

This definition is intended to cover a broad range of deceptive AI practices, while providing a clear framework for identifying and assessing instances of deception.

\section{Vulnerability Exploits are Scale Dependant}

The current version of the AI Act wants to prohibit an \emph{“AI system that exploits any of the vulnerabilities of a person or a specific group of persons.”} We argue that exploits are scale-dependent \cite{taleb2020}. Exploiting a person is different from exploiting a group of persons \cite{benn2022}. These differences should be further specified in the act. 

The exploitation of an individual, in the context of AI manipulation, can be conceptualised as the misuse of an AI system to leverage the inherent or contextual vulnerabilities of a person for undue gain, typically at the expense of the exploited individual. It will most often involve manipulative practices that subvert an individual's autonomy or privacy. As it is targeted at an individual, it will often involve a greater breach of the individual’s privacy, as the first step in exploiting an individual is to learn how an individual can be exploited. Although AI systems may possess some information rightfully, other information may be seen as an infringement of privacy rights. It may be based on intimate personal data that exploits psychological susceptibilities to alter individual behaviour. These practices can engender substantial harm and infringement of personal sovereignty, often unbeknownst to the individual in question.

To exploit a group is to identify a measurable factor in a group that has high predictive leverage, and thus targeting it results in a significant change to group behaviour. This exploitation results in a marked alteration in collective behaviour, often aligning with the exploitative entity's objectives at the cost of the group's autonomy and well-being. An example is the loss-avoidance tendency in most people, which can be used to manipulate risk-seeking behaviour within a group, thereby redirecting actions and decisions in a way that could potentially be harmful or against the group's inherent interests \cite{kahneman1979}. Furthermore, group exploitation can entail discriminatory practices, where AI systems unintentionally or deliberately target certain groups based on shared attributes such as race, gender, or socioeconomic status. This could lead to systemic biases, reinforcing social stereotypes, and perpetuating existing social inequities.

Considering the distinct characteristics and potential harms associated with individual and group exploitation by AI systems, different policy solutions may be required. Thus we recommend that it is crucial to differentiate between individual and group exploitation in the Act, in order to address the specific challenges present in each type of exploitation. For individual exploitation, regulations should emphasise stringent data protection and privacy standards, as well as mechanisms to ensure the individual's informed consent and autonomy in the face of persuasive AI technologies. As for group exploitation, the AI Act should incorporate measures to prevent systemic biases in AI systems and to guard against AI-induced social inequities and group-based discrimination. This may involve requirements for transparency and bias mitigation in the design, deployment, and auditing of AI systems.

\section{Defining Informed Decisions}

The EU AI Act seeks to regulate manipulation techniques that operate by \emph{"impairing the person’s ability to make an informed decision, thereby causing the person to take a decision that that person would not have otherwise taken."}

An informed decision, in its most basic form, refers to a choice made with full awareness and understanding of the relevant information, potential consequences, and the available alternatives. This concept plays a critical role in policy and regulation, particularly in contexts such as consent, consumer rights, medical decisions, and, as in this case, interactions with AI systems (e.g., \cite{eu2005}). Therefore, crafting a definition that is useful for policymakers and regulation necessitates an integration of principles from ethics, law, and cognitive sciences. In the context of the EU AI Act, an informed decision can be defined as:

A decision made by a person or group of persons, having full comprehension of the pertinent information, potential outcomes, and available alternatives, unimpaired by distorting external influences. This involves having clear, accurate, and sufficient information about the nature, purpose, and implications of the AI system in question, including but not limited to its functioning, data usage, potential risks, and the extent to which the system influences or informs their choices.

The facets of this definition are:

\begin{enumerate}
    \item Full comprehension: Decision-makers should understand the relevant information, potential implications, and available alternatives. In the context of AI, this may include an understanding of the nature of the AI system and its possible effects on their behaviour.
    \item Accurate and sufficient information: The information provided should be accurate, complete, and easy to understand. Any crucial information that could significantly influence decision-making should not be withheld or obscured.
    \item Absence of subliminal, manipulative, or deceptive techniques: The decision-making process should be free from covert or overt influences that could distort perception, judgment, or choice.
    \item Understanding of AI influence: Decision-makers should be made aware of the extent to which the AI system could be influencing their choices, allowing them to take this into account in their decision-making process.
\end{enumerate}

Incorporating this definition into the Act would provide a more robust framework for assessing whether an AI system has materially distorted a person's behaviour by impairing their ability to make an informed decision. It could guide the creation of standards for transparency and disclosure around AI systems, and underpin the development of regulatory measures aimed at protecting the rights and autonomy of individuals and groups interacting with AI.

\section{Defining Therapeutic Purposes}

In accordance with the EU AI act, prohibitions on manipulation \emph{‘shall not apply to AI systems intended to be used for approved therapeutical purposes on the basis of specific informed consent’.}

The excepting of ‘therapeutic’ applications of AI is well-reasoned in that we can envisage many systems, such as those used in psychiatric contexts, where manipulation or persuasion are crucial to therapeutic effectiveness. However, we believe this clause leaves an opportunity whereupon subliminal techniques can be employed if systems explicitly include therapeutic purposes in their terms of service. 

Therapeutic purposes are thus under-defined within the current draft of the EU AI Act with potentially hazardous implications. AI systems can often be dual-purpose technology \cite{feldstein2019}, in that the same innovation ostensibly marked as being for ‘therapeutic purposes’ can be used nefariously by malicious actors. We can envisage AI systems designed for uses such as mental healthcare or dermatology being used for psychometric targeting and biometric data harvesting respectively. A comparable situation is in place in pharmaceutical regulation, certain classes of drugs have both ‘therapeutic purposes’ and possible dual-uses (recreation, weaponization of doping). As such, a specific licensing regime for AI products and services marked as ‘therapeutic’ is needed.

Such a licensing regime would regulate AI therapies similarly to other software-based medical interventions under the umbrella of Digital Therapeutics (DTx). Digital therapeutics (DTx), are defined by the Digital Therapeutics Alliance as “evidence-based therapeutic interventions driven by high-quality software programs to prevent, manage, or treat a medical disorder or disease \cite{dtxalliance2023}”. However, even non-AI DTx face regulatory uncertainty within the EU. The EMA has no harmonised regulatory pathway for  DTx \cite{ema2020}. This is especially pertinent given the increasing skepticism towards unregulated AI therapy apps -  novel research shows that there is evidence that some of these can do harm \cite{akbar2020}. Most digital health apps are developed without medical professional involvement \cite{rassicruz2022}. 

Currently, few DTx products have completed conformity assessments, leading to limited regulatory precedents for risk classification. As regulatory bodies and DTx developers deepen their understanding of patient risks, clearer guidelines will emerge. Given the innovative nature of AI DTx products, it's imperative for notified bodies to maintain and expand their expertise for effective evaluations. There's a concern that a lack of expertise in these bodies could affect the quality of scientific reviews.

The decision to exclude 'therapeutic' AI from the EU AI Act's restrictions, leading to a poorly regulated market, poses significant risks to European citizens. Meanwhile, the regulation of AI digital therapeutics poses unique challenges, including increased cybersecurity and privacy demands for sensitive health data \cite{ema2021}. The EMA’s current post-authorisation management of medicines, including the Variation framework, must be adapted to accommodate updates to AI software linked to a medicinal product. 

A significant hurdle for the regulation of AI DTx is the absence of a standardised assessment framework. Currently, only Belgium and Germany have instituted DTx assessment frameworks at a national level and with significant incompatibilities presenting a barrier to European harmonisation \cite{efpia2023}. This results in ambiguous evidence requirements for developers, which may differ across markets. The diversity in expected evidence types further complicates this for AI digital therapeutics. Although randomised controlled trials have been employed for non-AI DTx, there's an anticipated shift towards real-world evidence in AI DTx evaluations \cite{gordon2018}. This shift stems from the belief that RCTs might not always be suitable, especially as DTx demands novel methods for ongoing real-world effectiveness assessments, given the adaptability and context-specificity of AI models \cite{yan2021}. Another challenge is that AI systems tend to generalise existing HTA frameworks, which evaluate products based on specific indications, necessitating distinct evidence and trials for each indication.

As per the current iteration of the EU AI Act, we recommend AI with ‘therapeutic purposes’ to refer specifically to AI digital therapeutics as regulated by the EMA. We however point out that this is insufficient given that digital therapeutics are underegulated at a European level. 

\section{Conclusion}

This paper attempts to clarify and define usable concepts in areas lacking in the EU AI Act Draft. To this end, we introduced new definitions and concepts in the following core areas - personality traits, preferences,  deceptive techniques, subliminal techniques, and purposefully manipulative techniques, group and individual vulnerabilities, informed decisions, and therapeutic purposes. When new concepts were introduced, it was to strengthen the understanding of currently present concepts, rather than to change the goals of the EU AI Act. Our recommendations are intended to provide a technical basis to legislators in formulating the EU AI Act. Our recommendations are targeted at ensuring the AI Act is precise and enforceable so that it may protect EU citizens from being subject to the deceptive and manipulative potential of current and future AI systems.

\section*{Acknowledgments}
We would like to thank Risto Uuk, Juan Pablo Bermúdez, Ariel Gil, and Lorenzo Pacchiardi for their valuable feedback.

\bibliographystyle{unsrt}  
\bibliography{references}

\end{document}